\def\year{2020}\relax
\documentclass[letterpaper]{article}
\usepackage{aaai20}
\usepackage{times}
\usepackage{helvet}
\usepackage{courier}
\usepackage{url}
\usepackage{graphicx}
\usepackage{booktabs}
\usepackage{amssymb}
\usepackage{amsmath}
\usepackage{times}
\usepackage{soul,color}
\usepackage{url}
\usepackage[hidelinks]{hyperref}
\usepackage[utf8]{inputenc}
\usepackage[small]{caption}
\usepackage{graphicx}
\usepackage{amsmath}
\usepackage{booktabs}
\usepackage{algorithm}
\usepackage{algorithmic}
\usepackage{amsmath}
\usepackage{stfloats}
\usepackage{amsfonts}
\usepackage{subfigure}

\title{Mega-Reward: Achieving Human-Level Play without Extrinsic Rewards}

\author{
Yuhang Song,\textsuperscript{\rm 1} 
Jianyi Wang,\textsuperscript{\rm 3} 
Thomas Lukasiewicz,\textsuperscript{\rm 1} 
Zhenghua Xu,\textsuperscript{\rm 1,2}\thanks{Corresponding author: Zhenghua Xu.} 
Shangtong Zhang,\textsuperscript{\rm 1}\\\Large 
\textbf{Andrzej Wojcicki},\textsuperscript{\rm 4} 
\textbf{Mai Xu}\textsuperscript{\rm 3} \\ 
\textsuperscript{\rm 1}Department of Computer Science, University of Oxford, United Kingdom\\
\textsuperscript{\rm 2}State Key Laboratory of Reliability and Intelligence of Electrical Equipment, Hebei University of Technology, China\\
\textsuperscript{\rm 3}School of Electronic and Information Engineering, Beihang University, China\\
\textsuperscript{\rm 4}Lighthouse\\
\{yuhang.song,thomas.lukasiewicz,shangtong.zhang\}@cs.ox.ac.uk, \{iceclearwjy,maixu\}@buaa.edu.cn \\ zhenghua.xu@hebut.edu.cn, andrzej@wojcicki.xyz
}

\begin{document}
\maketitle
\begin{abstract}
	Intrinsic rewards were introduced to simulate how human intelligence works; they are usually evaluated by intrinsically-motivated play, i.e., playing games without extrinsic rewards but evaluated with extrinsic rewards. However, none of the existing intrinsic reward approaches can achieve human-level performance under this very challenging setting of intrinsically-motivated play. In this work, we propose a novel megalomania-driven intrinsic reward (called \emph{mega-reward}), which, to our knowledge, is the first approach that achieves human-level performance in intrinsically-motivated play. Intuitively, mega-reward comes from the observation that infants' intelligence develops when they try to gain more control on entities in an environment; therefore, mega-reward aims to maximize the control capabilities of agents on given entities in a given environment. To formalize mega-reward, a relational transition model is proposed to bridge the gaps between direct and latent control. Experimental studies show that mega-reward (i) can greatly outperform all state-of-the-art intrinsic reward approaches, (ii) generally achieves the same level of performance as Ex-PPO and professional human-level scores, and (iii) has also a superior performance when it is incorporated with extrinsic rewards.
\end{abstract}

\section{Introduction}

Since humans can handle real-world problems without explicit extrinsic reward signals \cite{friston2010free}, intrinsic rewards \cite{oudeyer2009intrinsic} are introduced to simulate how human intelligence works.
Notable recent advances on intrinsic rewards include empowerment-driven \cite{klyubin2005all,klyubin2008keep,mohamed2015variational,montufar2016information}, count-based novelty-driven \cite{bellemare2016unifying,martin2017count,ostrovski2017count,tang2017exploration}, prediction-error-based novelty-driven \cite{achiam2017surprise,pathak2017curiosity,burda2018large,burda2018exploration}, stochasticity-driven \cite{florensa2017stochastic}, and diversity-driven \cite{SWLXX-AAAI-2019} approaches.
Intrinsic reward approaches are usually evaluated by \emph{intrinsically-motivated play}, where proposed approaches are used to play games without extrinsic rewards but evaluated with extrinsic rewards. However, though proved to be able to learn some useful knowledge \cite{florensa2017stochastic,SWLXX-AAAI-2019} or to conduct a better exploration \cite{burda2018large,burda2018exploration}, none of the state-of-the-art intrinsic reward approaches achieves a performance that is comparable to human professional players under this very challenging setting of intrinsically-motivated play.

In this work, we propose a novel megalomania-driven intrinsic reward (called \textit{mega-reward}), which, to our knowledge, is the first approach that achieves  human-level performance in intrinsically-motivated play. The idea of mega-reward originates from early psychology studies on \textit{contingency awareness} \cite{watson1966development,baeyens1990contingency,bellemare2012investigating}, where infants are found to have awareness of how entities in their observation are potentially under their control.
We notice that the way in which contingency awareness helps infants to develop their intelligence is to motivate them to have more control over the entities in the environment; therefore, we believe that having more control over the entities in the environment should be a very good intrinsic reward.
Mega-reward follows this intuition, aiming to maximize the control capabilities of agents on given entities in a given environment.

\begin{figure}
    \centering
	\includegraphics[width=1.0\columnwidth]{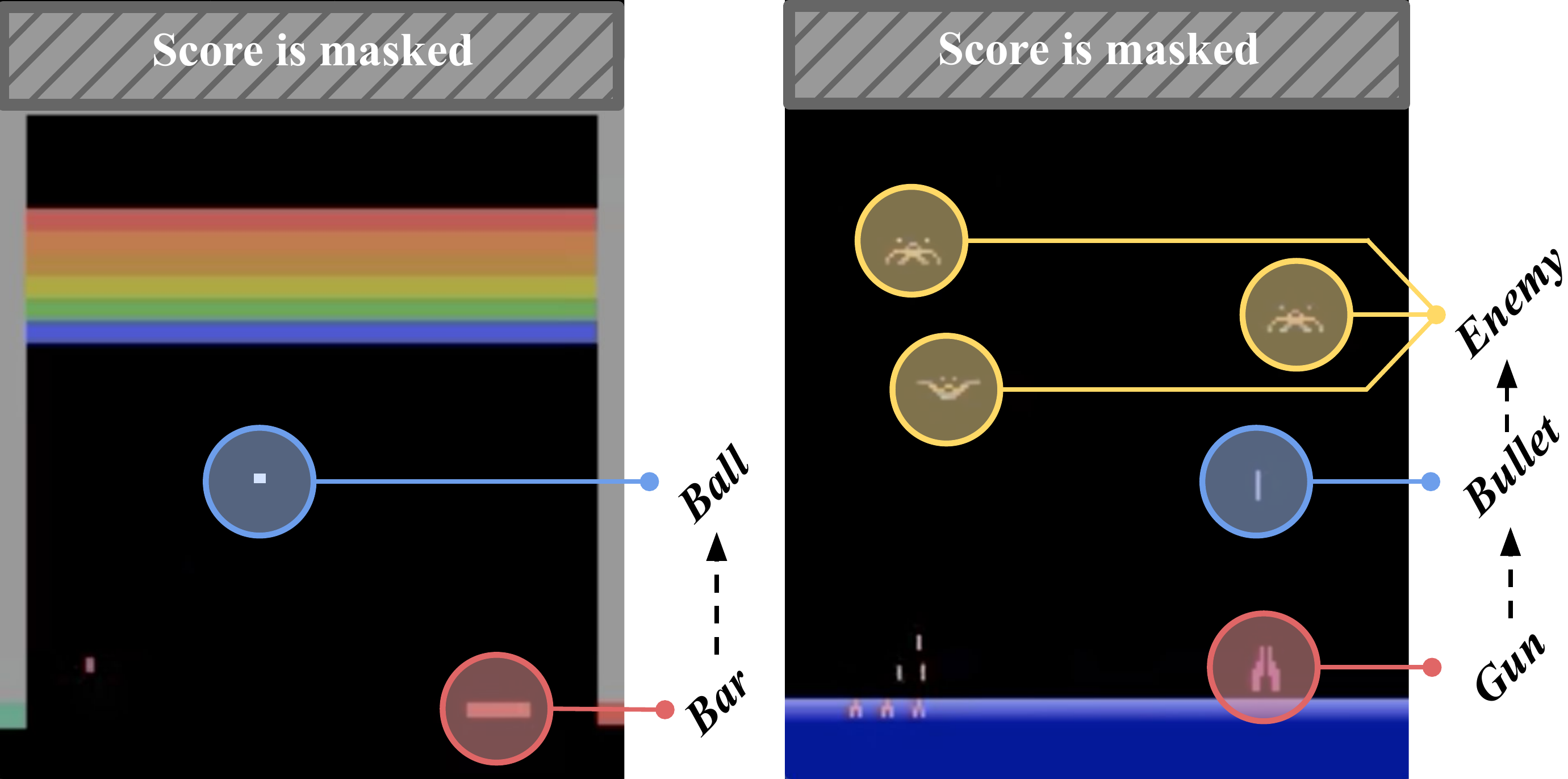}
	\caption{Latent control in \textit{Breakout} (left) and \textit{DemonAttack} (right).}
	\label{game-set}
\end{figure}

Specifically, taking the game \textit{Breakout} (shown in Fig.~\ref{game-set} (left)) as an example, if an infant is learning to play this game, contingency awareness may first motivate the infant to realize that he/she can control the movement of an entity, \textit{bar}; then, with the help of contingency awareness, he/she may continue to realize that blocking another entity, \textit{ball}, with the bar can result in the ball also being under his/her control. Thus, the infant's skills on playing this game is gradually developed by having more control on entities in this game.

Furthermore, we also note that entities can be controlled by two different modes: \emph{direct control} and \emph{latent control}. Direct control means that an entity can be controlled directly (e.g., \emph{bar} in \textit{Breakout}), while latent control means that an entity can only be controlled indirectly by controlling another entity (e.g., \emph{ball} is controlled indirectly by controlling \emph{bar}). In addition,  latent control usually forms a hierarchy in most of the games; the game \textit{DemonAttack} as shown in Fig.~\ref{game-set} (right) is an example: there is a \textit{gun}, which can be fired (direct control); then firing the gun controls \textit{bullets} (1st-level latent control); finally, the \textit{bullets} control \textit{enemies} if they eliminate enemies (2nd-level latent control).

Obviously, gradually discovering and utilizing the hierarchy of latent control helps infants to develop their skills on such games.
Consequently, mega-reward should be formalized by maximizing not only direct control, but also latent control on entities. This thus requests the formalization of both direct and latent control. However, although we can model direct control with an attentive dynamic model \cite{choi2018contingency}, there is no existing solution that can be used to formalize latent control. Therefore, we further propose a \textit{relational transition model} (RTM) to bridge the gap between direct and latent control by learning how the transition of each entity is related to itself and other entities. For example, the agent's direct control on entity $A$ can be passed to entity $B$ as latent control if $A$ implies the transition of $B$. With the help of RTM, we are able to formalize mega-reward, which is computationally tractable.

Extensive experimental studies have been conducted on 18 Atari games and the ``{\em noisy TV}'' domain~\cite{burda2018large}; the experimental results show that (i) mega-reward significantly outperforms all six state-of-the-art intrinsic reward approaches, (ii) even under the very challenging setting of intrinsically-motivated play, mega-reward (without extrinsic rewards) still achieves generally the same level of performance as two benchmarks (with extrinsic rewards), Ex-PPO and professional human-level scores, and (iii) the performance of mega-reward is also superior when it is incorporated with extrinsic rewards, outperforming state-of-the-art approaches in two different settings.

This paper's contributions are briefly as follows:
(1) We propose a novel intrinsic reward, called mega-reward, which aims to maximize the control capabilities of agents on given entities in a given environment.
(2) To realize mega-reward, we further propose a relational transition model (RTM)  to bridge the gap between direct and latent control.
(3) Experiments on 18 Atari games and the ``{\em noisy TV}'' domain show that mega-reward (i) greatly outperforms all state-of-the-art intrinsic reward approaches, (ii)~generally achieves the same level of performance as two benchmarks, Ex-PPO and professional human-level scores, and (iii)~also has a superior performance when it is incorporated with extrinsic rewards.
Easy-to-run code is released in https://github.com/YuhangSong/Mega-Reward.

\begin{figure*}
	\centering
	\includegraphics[width=0.6\textwidth]{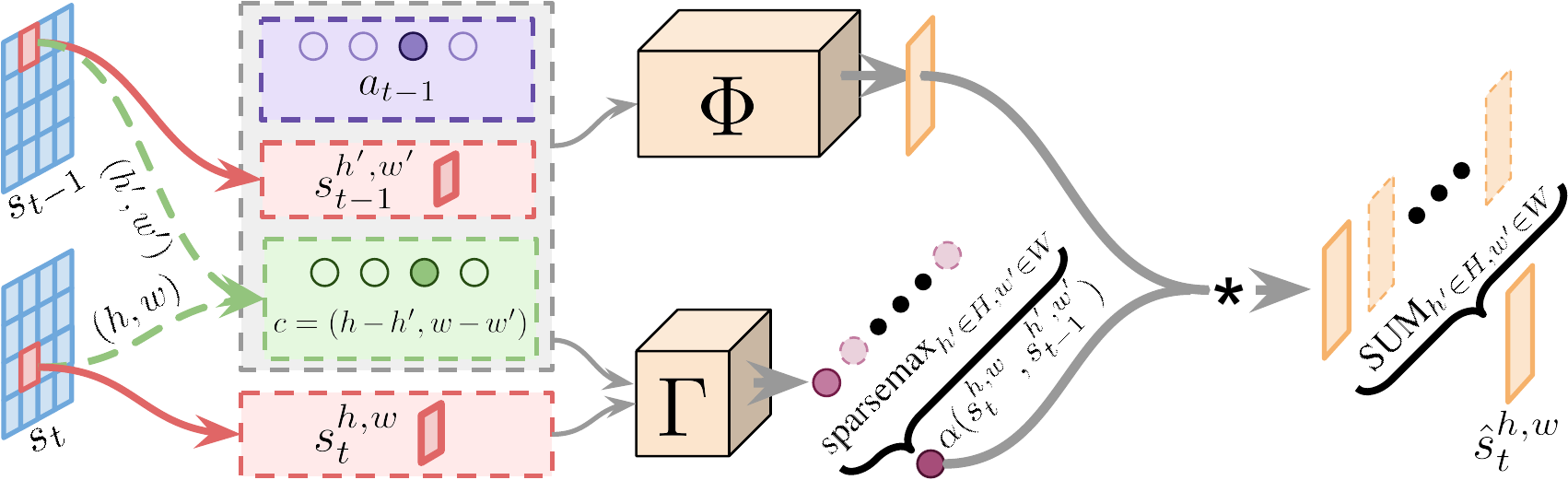}
	\caption{Relational transition model.}
	\label{fig-rtm}
\end{figure*}

\section{Direct Control}

We start with the notion of \textit{direct control}.
Generally, we consider the effect of the action $a_{t-1} \in \mathcal{A}$ on the state $s_{t} \in \mathcal{S}$ as direct control.
In practice, we are more interested in how different parts of a visual state are being directly controlled by $a_{t-1}$.
Thus, prevailing frameworks \cite{jaderberg2016reinforcement,choi2018contingency} mesh $s_{t}$ into subimages, as shown in Fig.~\ref{fig-rtm}, where we denote a subimage of $s_{t}$ at the coordinates $(h,w)$ as $s^{h,w}_{t} \in \mathcal{S}^{H,W}$.
The number of possible coordinates $(h, w)$ and space of the subimage $\mathcal{S}^{H,W}$ is determined by the granularity of the meshing $H,W$.
Then, we can define the quantification of how likely each $s^{h,w}_{t}$ is being directly controlled by $a_{t-1}$ as $\alpha(s^{h,w}_{t},a_{t-1}) \in \mathbb{R}$.

The state-of-the-art method \cite{choi2018contingency} models $\alpha(s^{h,w}_{t},a_{t-1})$ with an \textit{attentive dynamic model (ADM)}, 
which predicts $a_{t-1}$ from two consecutive states $s_{t-1}$ and $s_{t}$.
The key intuition is that ADM should attend to the most relevant part of the states $s_{t-1}$ and $s_{t}$, which is controllable by $a_{t-1}$, to be able to classify $a_{t-1}$.
Thus, a \textit{spatial attention mechanism} \cite{bahdanau2014neural,xu2015show} can be applied to ADM to model $\alpha(s^{h,w}_{t},a_{t-1})$:
\begin{equation}
  \label{adm-1}
  e^{h,w}_{t} = \Theta \left( \left[ s^{h,w}_{t}-s^{h,w}_{t-1} ; s^{h,w}_{t} \right] \right) \in \mathbb{R}^{|\mathcal{A}|}
\end{equation}
\begin{equation}
  \label{adm-2}
  \bar{\alpha}(s^{h,w}_{t}, a_{t-1}) = \Lambda \left( s^{h,w}_{t} \right) \in \mathbb{R}
\end{equation}
\begin{equation}
  \label{adm-3}
  \alpha(s^{h,w}_{t},a_{t-1}) = \textrm{sparsemax} \left( \bar{\alpha}(s^{h,w}_{t}, a_{t-1}) \right) \in \mathbb{R}
\end{equation}
\begin{equation}
  \label{adm-4}
  \!\!p( \hat{a}_{t-1} | s_{t-1}, s_{t} ) =  \textrm{SoM}\! \left( \!\sum_{\substack{h' \in H, w' \in W}\!\!\!\!\!\!\!\!\!\!\!\!\!\!\!\!\!\!\!\!\!\!\!\!} \alpha(s^{h,w}_{t},a_{t-1}) \cdot e^{h,w}_{t}\! \right), 
\end{equation}
where $\Theta$ and $\Lambda$ are two parameterized models, $e^{h,w}_{t}$ is the logits of the probability of the predicted action $p( \hat{a}_{t-1} | s_{t-1}, s_{t} )$ before masking it by the spatial attention $\alpha(s^{h,w}_{t},a_{t-1})$, $\textrm{SoM}$ is the softmax operation, and $\bar{\alpha}(s^{h,w}_{t}, a_{t-1})$ is the spatial attention mask before converting it into a probability distribution $\alpha(s^{h,w}_{t},a_{t-1})$ using the sparsemax operator \cite{martins2016softmax}.
The models can be optimized with the standard cross-entropy loss $\mathcal{L}_\textrm{action}(a_{t-1}, \hat{a}_{t-1})$ relative to the ground-truth action $a_{t-1}$ that the agent actually has taken.
More details, including additional \textit{attention entropy regularization losses}, can be found in \cite{choi2018contingency}.

\section{From Direct Control to Latent Control}

Built on the modelled direct control $\alpha(s^{h,w}_{t},a_{t-1})$, we propose to study the notion of \textit{latent control}, which means the effect of $a_{t-n}$ on the state $s_{t}$, where $n>1$.
Like in direct control, we are interested in how different parts of a visual state $s^{h,w}_{t}$ are being latently controlled, thus, we define the quantification of how likely $s^{h,w}_{t}$ is being latently controlled by $a_{t-n}$ as $\alpha(s^{h,w}_{t},a_{t-n})$.
As an example for latent control quantification $\alpha(s^{h,w}_{t},a_{t-n})$,  consider the game \textit{DemonAttack} in Fig.~\ref{game-set} (right).
In this case,  consider $a_{t-n}$ to be the action at step $t-n$ that shoots a bullet from the \textit{gun}, $s^{h,w}_{t}$ to be the subimage containing one of the \textit{enemy} at step $t$.
Clearly, $s^{h,w}_{t}$ is influenced by $a_{t-n}$.
But this influence needs $n$ steps to take effect, where $n$ is unknown; also, it involves an intermediate entity \textit{bullet}, i.e., the \textit{gun} ``controls'' the \textit{bullet}, the \textit{bullet} ``controls'' the \textit{enemy}.
Due to the nature of delayed influence and involvement of intermediate entities, we call it latent control, in contrast to direct control.

To compute $\alpha(s^{h,w}_{t},a_{t-n})$, one possible way is to extend ADM to longer interval, i.e., a variant of ADM takes in $s_{t-n}$, $s_{t}$ and makes predictions of a sequence of actions $a_{t-n}, a_{t-n+1}, ..., a_{t-1}$.
However, $n$ is not known in advance, and we may need to enumerate over $n$.
We propose an alternative solution, based on the observation that latent control always involves some intermediate entities.
For example, the latent control from the \textit{gun} to the \textit{enemy} in \textit{DemonAttack} in Fig.~\ref{game-set} (right) involves an intermediate entity \textit{bullet}.
Thus, how likely the \textit{enemy} is latently controlled can be quantified if we can model that (1) an action directly controls the \textit{gun}, (2)~the \textit{gun} directly controls the \textit{bullet}, and (3) the \textit{bullet} directly controls the \textit{enemy}.
In this solution, latent control is broken down into several direct controls, which avoids dealing with the unknown $n$.
As can be seen, this solution requires modelling not only the direct control of an action on a state, but also the direct control of a state on the following state, which is a new problem.
Formally speaking, we first quantify how likely $s^{h,w}_{t}$ is controlled by $s^{h',w'}_{t-1}$ with $\alpha(s^{h,w}_{t}, s^{h',w'}_{t-1}) \in \mathbb{R}$.
Then, we can formally express the above idea by
\begin{equation}
\label{C-C-model}
  \alpha(s^{h,w}_{t},a_{t-n}) = \sum_{h' \in H,w' \in W\!\!\!\!\!\!\!\!\!\!\!\!\!\!\!\!\!\!\!} \alpha(s^{h,w}_{t}, s^{h',w'}_{t-1}) \alpha(s^{h',w'}_{t-1}, a_{t-n}).
\end{equation}
Thus, Eq.\ \eqref{C-C-model} derives $\alpha(s^{h,w}_{t},a_{t-n}) $ from $\alpha(s^{h',w'}_{t-1}, a_{t-n})$.
If we keep applying Eq.\ \eqref{C-C-model} on $\alpha(s^{h',w'}_{t-1}, a_{t-n})$, we can eventually derive it from $\alpha(s^{h',w'}_{t-n+1}, a_{t-n})$, which is the qualification of direct control defined in the last section and can be computed via Eqs.~\eqref{adm-1} to \eqref{adm-4}.
Thus, $\alpha(s^{h,w}_{t},a_{t-n})$ can be computed as long as we know $\alpha(s^{h,w}_{t'}, s^{h',w'}_{t'-1})$ for all $t' \in \left[ t-n+2, t \right]$.
That is, $\alpha(s^{h,w}_{t}, s^{h',w'}_{t-1})$ bridges the gap between direct and latent control.
Furthermore, since $\alpha(s^{h,w}_{t}, s^{h',w'}_{t-1})$ models how a part of the previous state $s^{h',w'}_{t-1}$ implies a part of the current state $s^{h,w}_{t}$, it reveals the need of a new form of transition model, which contains information about the relationships between different parts of the state underlying the transition of full states.
Thus, we call it a \textit{relational transition model (RTM)}.
In the next section, we introduce our method to learn RTM efficiently.

\section{Relational Transition Model}

To produce an approximation of $\alpha(s^{h,w}_{t}, s^{h',w'}_{t-1})$, we propose relational transition models (RTMs), the general idea behind which is introducing a spatial attention mechanism to the \textit{transition model}.
Specifically, Fig.~\ref{fig-rtm} shows the structure of an RTM, which consists of two parameterized models, namely, $\Phi$ for relational transition modeling and $\Gamma$ for attention mask estimation.
We first define the forward function of $\Phi$; it makes a prediction of the transition from~$s^{h',w'}_{t-1}$ to~$s^{h,w}_{t}$:
\begin{equation}
	\label{rtm-forward}
  \hat{s}^{h,w}_{t} = \sum_{\substack{h' \in H, w' \in W}\!\!\!\!\!\!\!\!\!\!\!\!\!\!\!\!\!\!\!\!\!} \alpha(s^{h,w}_{t}, s^{h',w'}_{t-1}) \Phi \left(\left[s^{h',w'}_{t-1},a_{t-1},c\right]\right).
\end{equation}
Here, $\hat{s}^{h,w}_{t}$ represents the prediction of $s^{h,w}_{t}$.
Also, note that apart from taking in $s^{h',w'}_{t-1}$, $\Phi$ also takes in the relative coordinates $c=(h-h',w-w')$ and $a_{t-1}$, both as one-hot vectors, so that the model $\Phi$ knows the relative position of the part to predict and the action taken.
Furthermore, $\alpha(s^{h,w}_{t}, s^{h',w'}_{t-1})$ is the estimated attention mask of predicting $s^{h,w}_{t}$ from $s^{h',w'}_{t-1}$, which models how informative each $s^{h',w'}_{t-1}$ of different ${h',w'}$ is for the prediction of $s^{h,w}_{t}$, i.e., how likely $s^{h',w'}_{t-1}$ controls $s^{h,w}_{t}$.
$\alpha(s^{h,w}_{t}, s^{h',w'}_{t-1})$ is estimated by the model~$\Gamma$.
Specifically, $\Gamma$ first estimates $\bar{\alpha}(s^{h,w}_{t}, s^{h',w'}_{t-1})$ via
\begin{equation}
	\label{model-Gamma}
	\bar{\alpha}(s^{h,w}_{t}, s^{h',w'}_{t-1}) = \Gamma\left(\left[s^{h,w}_{t},s^{h',w'}_{t-1},a_{t-1},c \right]\right)\,,
\end{equation}
which is later sparsemaxed over $h' \in H, w' \in W$ to compute 
\begin{equation}
	\label{gamma-softmax}
	\alpha(s^{h,w}_{t}, s^{h',w'}_{t-1}) = \textrm{sparsemax} \left(\bar{\alpha}(s^{h,w}_{t}, s^{h',w'}_{t-1})\right)\,.
\end{equation}
We train RTM end-to-end with $L_\textrm{transition} = \textrm{MSE}(\hat{s}^{h,w}_{t},s^{h,w}_{t})$.
As an intuitive explaination of RTM, taking the game \textit{Breakout} (shown in Fig.~\ref{game-set} (left)) as an example, $\Phi$ makes three predictions of the current ball based on the previous ball, bar, and brick.
Since the final prediction of the current ball is the weighted combination of these three predictions, $\Gamma$ is further used to estimate the weights of this combination, measuring different control effects that the previous ball, bar, and brick have on the current ball.
We thus propose $\Phi$ and $\Gamma$ as relational transition models.

RTM has introduced separated forwards over every $h'\,{ \in}\, H$, $w' \,{\in}\, W$, $h \in H$, and $w\,{ \in}\, W$; however, by putting the separated forwards into the batch axis, the computing is well parallelized.
We reported the running times and included code in
the extended paper \cite{supplementary_material}.

\section{Formalizing Intrinsic Rewards}

Summarizing the previous sections, ADM and RTM model $\alpha(s^{h',w'}_{t-n+1}, a_{t-n})$ and $\alpha(s^{h,w}_{t}, s^{h',w'}_{t-1})$, respectively. Based on this,
$\alpha(s^{h,w}_{t},a_{t-n})$ can be modelled via Eq.~\eqref{C-C-model}.
In this section, we formalize the intrinsic reward from $\alpha(s^{h,w}_{t},a_{t-n})$.

First, $\{\alpha(s^{h,w}_{t},a_{t-n})\}^{n\in\left[ 1,t \right]}$ contains all the information about what is being controlled by the agent in the current state, considering all the historical actions with both direct and latent control.
Clearly, computing all components in the above set is intractable as $t$ increases.
Thus, we define a quantification of accumulated latent control $g^{h,w}_{t} \in \mathbb{R}$, which is a discounted sum of  $\alpha(s^{h,w}_{t},a_{t-n})$ over~$n$:
\begin{equation}
\label{G}
g^{h,w}_{t} = \sum_{n\in\left[ 1,t \right]} \rho^{n-1} \alpha(s^{h,w}_{t},a_{t-n}),
\end{equation}
where $\rho$ is a discount factor, making $\alpha(s^{h,w}_{t},a_{t-n})$ with $n>\!>1$ have a lower contribution to the estimation of $g^{h,w}_{t}$.
Then, we show that $g^{h,w}_{t}$ can be computed from $g^{h,w}_{t-1}$ and $\alpha(s^{h,w}_{t},a_{t-1})$ without enumerating over $n$ (see proof of Lemma 1 in the extended paper \cite{supplementary_material}):
\begin{equation}
\label{w-iteration}
g^{h,w}_{t} = \rho \sum_{\substack{h' \in H, w' \in W}\!\!\!\!\!\!\!\!\!\!\!\!\!\!\!\!\! \!\!\!} \alpha(s^{h,w}_{t}, s^{h',w'}_{t-1}) g^{h',w'}_{t-1} + \alpha(s^{h,w}_{t},a_{t-1}),
\end{equation}
which reveals that we can maintain an $H \times W$ memory for $g^{h,w}$, and then update $g^{h,w}_{t-1}$ to $g^{h,w}_{t}$ at each step with $\alpha(s^{h,w}_{t}, s^{h',w'}_{t-1})$ and $\alpha(s^{h,w}_{t},a_{t-1})$ according to \eqref{w-iteration}.
The intuitive integration of $g^{h,w}_{t}$ is an overall estimation of what is being controlled currently, both directly and latently, considering the effect of all historical actions.
This also coincides with the intuition that humans do not explicitly know what they latently control for each historical action.
Instead, we maintain an overall estimation of what is under the historical actions' control, both directly and latently.
At last, to maximize $\sum_{\substack{h \in H, w \in W}} g^{h,w}_{t=T}$, where $T$ is the terminal step, the intrinsic reward (our mega-reward) at each step $t$ {should~be}:
\begin{equation}
\label{r-in-delta}
\mbox{$r^{\textrm{meg}}_t = \sum_{\substack{h \in H, w \in W}} \left( g^{h,w}_{t} - g^{h,w}_{t-1} \right)\,.$}
\end{equation}

\section{Experiments}

In extensive experiments, we evaluated the performance of mega-reward. We first report on the evaluation on 18 Atari games under the very challenging settings of intrinsically-motivated play, where a case study is used to visualize how each part of mega-reward works, and mega-reward is compared with six state-of-the-art intrinsic rewards, the benchmark of a PPO agent with access to extrinsic rewards (\emph{Ex-PPO}), and the benchmark of professional human-level scores, to show its superior performance. Then, we further investigate two possible ways to integrate mega-reward with extrinsic rewards. Finally, a few failure cases of mega-reward are studied, showing possible topics for future research.

Mega-reward is implemented on PPO in \cite{schulman2017proximal} with the same set of hyper-parameters, along with $H \times W=4 \times 4$ and $\rho=0.99$.
$H \times W=4 \times 4$ is a trade-off between efficiency and accuracy.
An ablation study on value settings of $H \times W$ over the game \textit{Breakout} is available in the extended paper \cite{supplementary_material}, showing that $4 \times 4$ is sufficient to achieve a reasonable accuracy, while having the best efficiency.
The network structures of $\Phi$ and $\Gamma$ are provided in the extended paper \cite{supplementary_material}.
The hyper-parameters of the other baseline methods are set as in the corresponding original papers.
The environment is wrapped as in \cite{burda2018large,mnih2015human}.

Due to the page limit, running times, additional ablation studies (e.g., of components in mega-reward), and additional comparisons under other settings (e.g., the setting when agents have access to both intrinsic and extrinsic rewards) are provided in the extended paper \cite{supplementary_material}.

\subsection{Intrinsically-Motivated Play of Mega-Reward}
\label{intrinsically-motivated-play-with-mega-reward}

Intrinsically-motivated play is an evaluation setting where the agents are trained by intrinsic rewards only, and the performance is evaluated using extrinsic rewards.
To make sure that the agent cannot gain extra information about extrinsic rewards,  the displayed score in each game is masked out.
To ensure a fair comparison, all baselines are also provided with a feature map $g^{h,w}_{t}$ as an additional channel.
Here, all agents are run for 80M steps, with the last $50$ episodes averaged as the final scores and reported in Table~\ref{raw-score}.
The evaluation is conducted over 18 Atari games.

\paragraph{Case Study.}

\begin{table}
	\caption{Comparison of mega-reward against six baselines.}
	\resizebox{\columnwidth}{!}{
		\begin{tabular}{lccccccc}
			Game                  & \textit{Emp} & \textit{Cur} & \textit{RND} & \textit{Sto} & \textit{Div} &  \textit{Dir} & \textit{Meg}\\
			\midrule
			\textit{Seaquest}    & 612.2                       & 422.2                       & 324.2          & 103.5                           & 129.2                       & 323.1                  & \textbf{645.2}                 \\
			\textit{Bowling}      & 103.4                       & \textbf{156.2}                       & 77.23         & 86.23                           & 79.21                       & 113.3                  & 82.72                 \\
			\textit{Venture}      & 62.34                      & 0.0                       & 83.12         & 61.32                           & 95.67                       & 86.21                  & \textbf{116.6}                \\
			\textit{WizardOfWor}  & 526.2                       & 562.3                      & 702.5         & 227.1                          & 263.1                       & 723.7                  & \textbf{1030}                 \\
			\textit{Asterix}      & 1536                       & 1003                      & 462.3        & 304.2                          & 345.6                       & 1823                  & \textbf{2520}                 \\
			\textit{Robotank}     & \textbf{5.369}                       & 3.518                      & 3.619         & 4.164                           & 2.639                       & 1.422                  & 2.310                 \\
			\textit{BeamRider}    & 944.1                       & 864.2                     & 516.3         & 352.1                           & 381.2                      & 1273                  & \textbf{1363}                 \\
			\textit{BattleZone}   & 3637                       & 4625                      & \textbf{8313}         & 0.0                           & 0.0                       & 2262                  & 3514                \\
			\textit{KungFuMaster} & 424.9                       & \textbf{3042}                       & 652.1        & 245.1                           & 523.9                      & 423.7                  & 352.4                 \\
			\textit{Centipede}    & 1572                       & 3262                     & \textbf{4275}         & 1832                           & 1357                       & 2034                  & 2001                 \\
			\textit{Pong}         & -7.234                       & -8.234                     & -17.42        & -16.52                           & -14.53                      & -17.62                  & \textbf{-3.290}                 \\
			\textit{AirRaid}      & 1484                       & 1252                      & 942.3          & 723.4                           & 1426                      & 1583                  & \textbf{2112}                 \\
			\textit{DoubleDunk}   & -18.26                       & -20.42                     & -17.34         & -19.34                           & -18.35                       & -17.72                  & \textbf{-13.58}                 \\
			\textit{DemonAttack}  & 9259                       & 69.14                      & 412.4          & 57.14                          & 90.23                       & 7838                  & \textbf{10294}                \\
			\textit{Berzerk}      & 735.7                       & 363.1                      & 462.4       & 157.2                           & 185.2                       & 413.3                  & \textbf{764.6}                 \\
			\textit{Breakout}     & 201.4                       & 145.3                      & 125.5         & 113.5                           & 1.352                       & 125.2                  & \textbf{225.3}                 \\
			\textit{Jamesbond}    & 523.2                       & 603.0                      & 201.2        & 0.0                           & 0.0                       & 1383                  & \textbf{3223}                 \\
			\textit{UpNDown}      & 8358                       & 8002                      & 2352          & 331.3                           & 463.3                       & 60528                  & \textbf{124423}
		\end{tabular}}
	\label{raw-score}
\end{table}

\begin{figure*}
	\centering
	\includegraphics[width=0.9\textwidth]{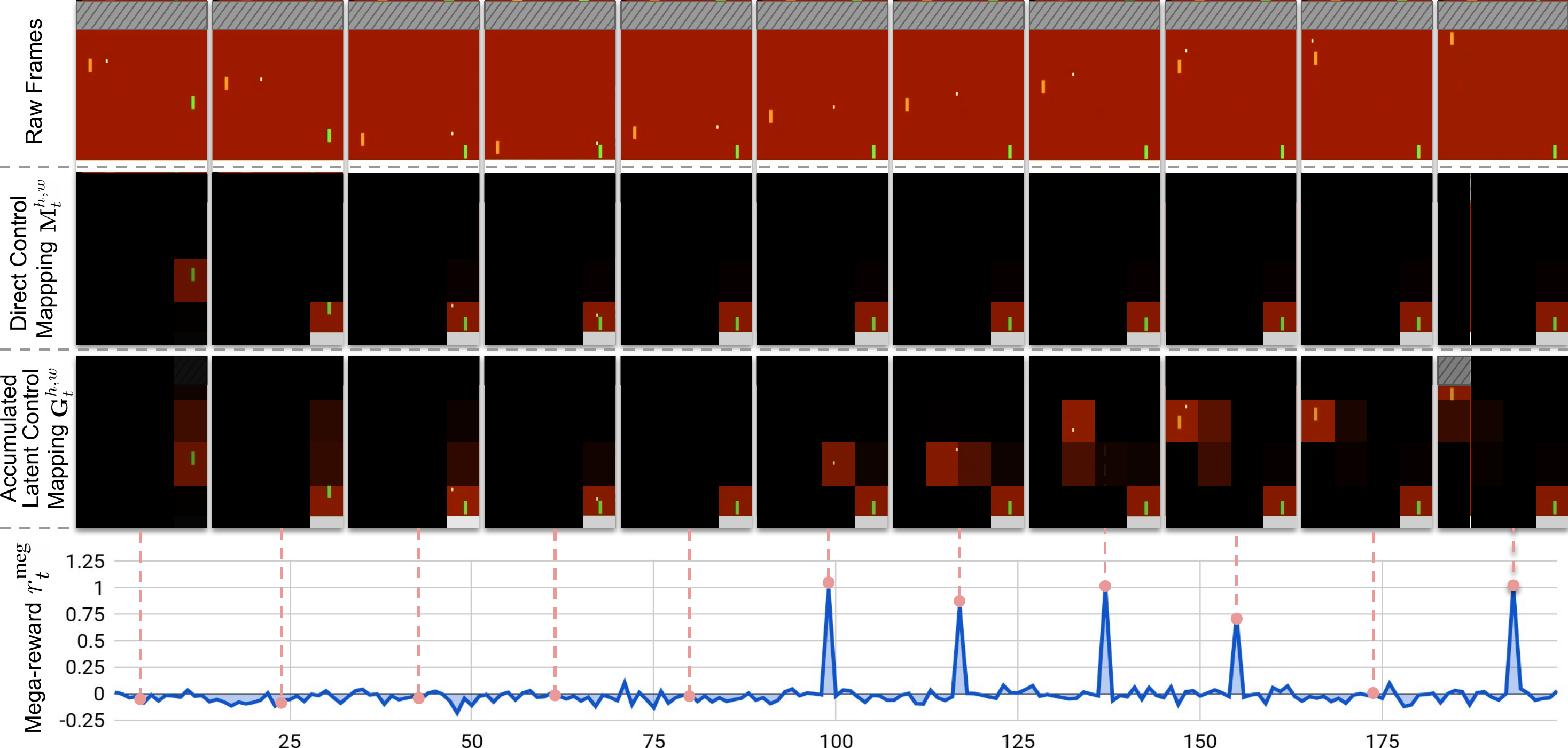}
	\caption{Case study: the example of \textit{Pong}.}
	\label{fig1}
\end{figure*}

Fig. \ref{fig1} visualizes  how each component in our method works as expected.
The 1st row is a frame sequence.
The 2nd row is the corresponding direct control map $\alpha(s^{h,w}_{t},a_{t-1})$, indicating how likely each grid being directly controlled by $a_{t-1}$. As expected, the learned map shows the grid containing the bar  being directly controlled.
The 3rd row is the accumulated latent control map $g^{h,w}_{t}$, indicating how likely each grid being controlled (both directly and latently) by historical actions.
As expected, the learned map shows:
(1) only the bar is under control before the bar hits the ball (frames 1--5);
(2) both the bar and the ball are under control after the bar has hit the ball (frames 6--10); and
(3) the bar, ball, and displayed score are all under control if the opponent missed the ball (frame 11).
The 4th row is mega-reward $r^{\textrm{meg}}_t$, obtained by Eq. \eqref{r-in-delta} from the map in the 3rd row.
As expected, it is high when the agent controls a new grid in the 3rd row (achieving more control over the grids in the state).

\paragraph{Against Other Intrinsic Rewards.}

To show the superior performance of mega-reward (denoted \textit{Meg}), we first compare its performance with those of six state-of-the-art intrinsic rewards, i.e., \textit{empowerment-driven} (denoted \textit{Emp})~\cite{mohamed2015variational}, \textit{curiosity-driven} (denoted \textit{Cur})~\cite{burda2018large}, \textit{RND}~\cite{burda2018exploration}, \textit{stochasticity-driven} (denoted \textit{Sto})~\cite{florensa2017stochastic}, \textit{diversity-driven} (denoted \textit{Div})~\cite{SWLXX-AAAI-2019}, and a mega-reward variant with only direct control (denoted \textit{Dir}).
Results for more baselines can be found in the extended paper \cite{supplementary_material}.
By the experimental results in Table~\ref{raw-score}, mega-reward outperforms all six baselines substantially. In addition, we also have the following findings: (i) \textit{Sto} and \textit{Div}  are designed for games with explicit hierarchical structures, so applying them on Atari games with no obvious temporal hierarchical structure will result in the worst performance among all baselines. (ii) \textit{Dir} is also much worse than the other baselines, proving the necessity of latent control in the formalization of mega-reward. (iii) The failure of the empowerment-driven approach states that applying information theory objectives to complex video games like Atari ones is an open problem.
A detailed discussion of the benefits of mega-reward over other intrinsically motivated approaches can be found in the extended paper \cite{supplementary_material}.
Videos demonstrating the benefits of mega-reward on all 57 Atari games can be found in the released code \cite{code}.

\paragraph{Against Two Benchmarks.}

\begin{figure}
	\centering
	\includegraphics[width=1.0\columnwidth]{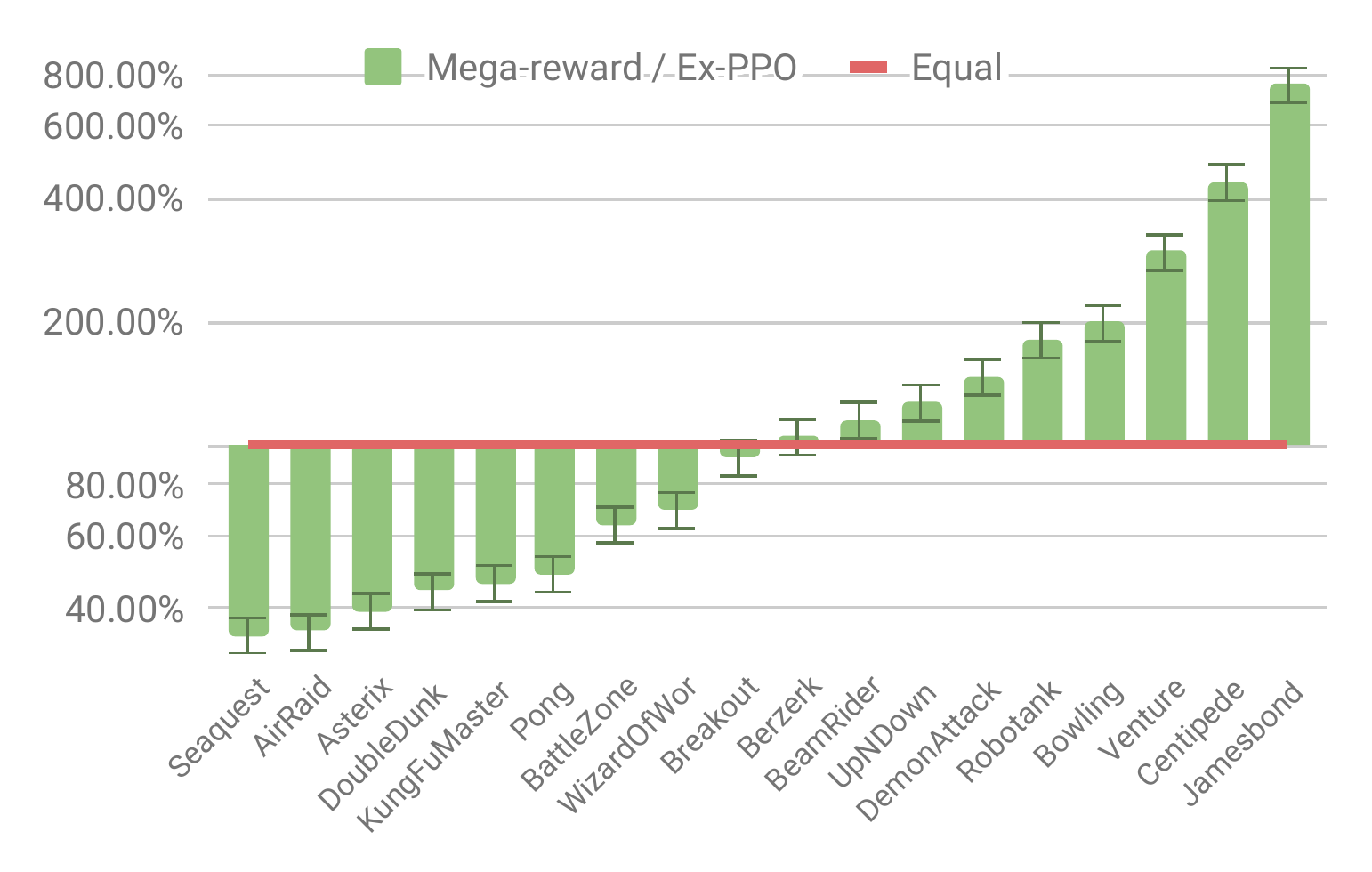}
	\caption{Mega-reward against the benchmark of Ex-PPO.}
	\label{mega-ex-ppo}
\end{figure}

In general, the purpose of evaluating intrinsic rewards in intrinsically-motivated play is to investigate if the proposed intrinsic reward approaches can achieve the same level of performance as two benchmarks: PPO agents with access to extrinsic rewards (denoted \emph{Ex-PPO}) and professional human players. Therefore, we evaluate mega-reward using a relative score against two such benchmarks, which can be formally defined as
\begin{equation}
\label{relative-score}
S_{\textrm{Relative}} = \frac{S_{\textrm{Mega-reward}}-S_{\textrm{Random}}}{S_{\textrm{Benchmark}}-S_{\textrm{Random}}} \times 100\%,
\end{equation}
where $S_{\textrm{Relative}}$ $>$ $100\%$ means that mega-reward achieves a better performance than the corresponding benchmark, $S_{\textrm{Relative}}$ $<$ $100\%$ that it achieves a worse performance, and $S_{\textrm{Relative}}$ $=$ $0\%$ is random play.

Fig.~\ref{mega-ex-ppo} shows the comparative performance of mega-reward against Ex-PPO on 18 Atari games, where mega-reward greatly outperforms the Ex-PPO benchmark in 8 games, and is close to the benchmark in 2 games. These results show that mega-reward generally achieves  the same level of or a comparable performance as Ex-PPO (though strong on some games and weak on others); thus, the proposed mega-reward is as informative as the human-engineered extrinsic rewards.

\begin{figure}
	\centering
	\includegraphics[width=1.0\columnwidth]{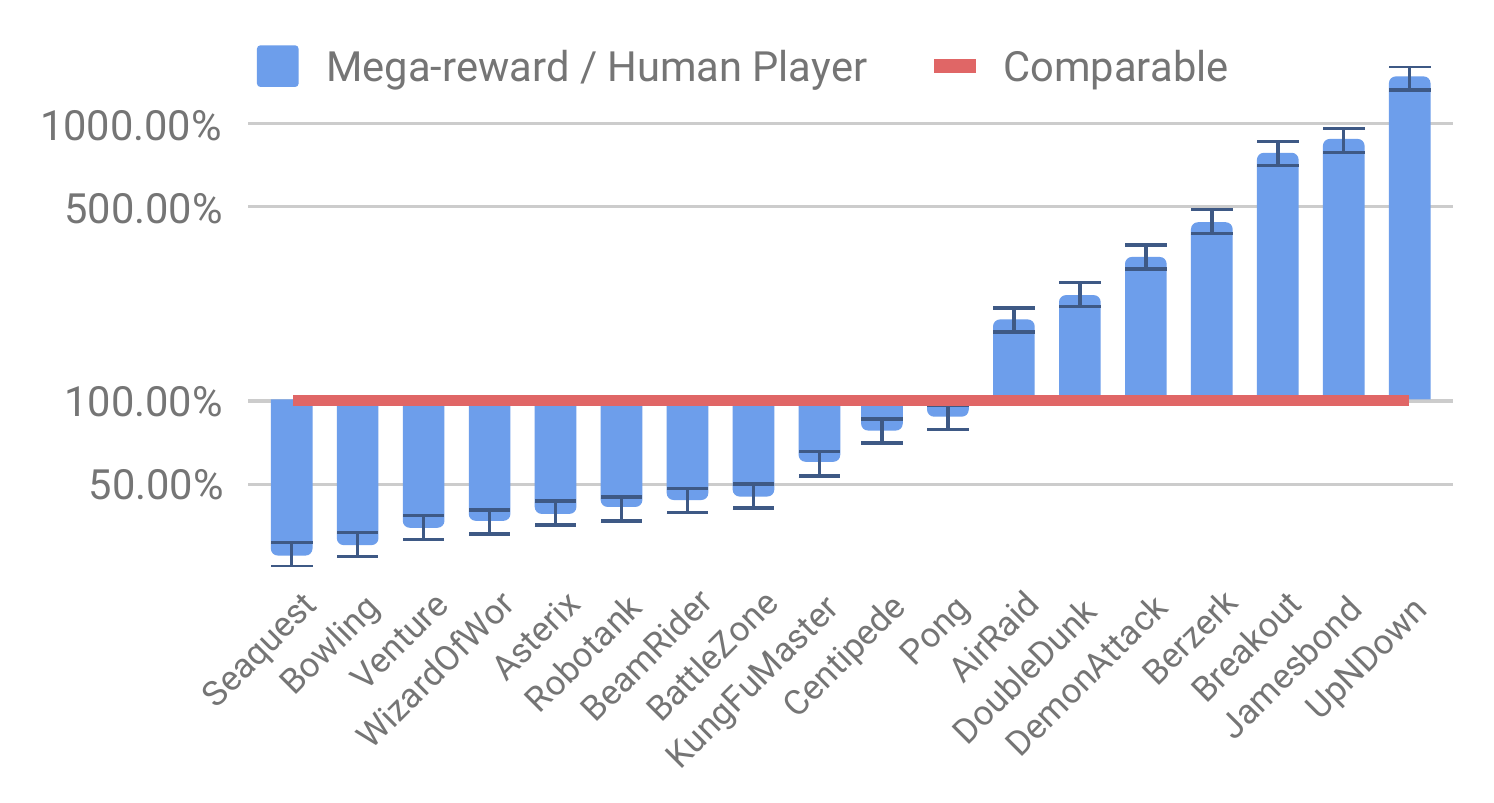}
	\caption{Mega-reward against the benchmark of human player.}
	\label{mega-human}
\end{figure}

Similarly, Fig.~\ref{mega-human} shows the  comparative performance of mega-reward against professional human players. As the performance of professional human players (i.e., professional human-player scores) on 16 out of 18 Atari games have already been measured by \cite{mnih2015human}, we measure the professional human-player scores on \textit{AirRaid} and \textit{Berzerk} using the same protocol. Generally, in Fig.~\ref{mega-human}, mega-reward greatly outperforms the professional human-player benchmark in 7 games, and is close to the benchmark in 2 games. As the professional players are equipped with strong prior knowledge about the game and the scores displayed in the state, they show a relatively high-level of human skills on the corresponding games. Thus, the results sufficiently prove that mega-reward has generally reached the same level of (or a comparable) performance as a human player.

\subsection{Pretraining with Mega-Reward}
\label{pretrain-with-mega-reward}

In many real-world cases, the agent may have access to the dynamics of the environment before the extrinsic rewards are available \cite{Ha2018WorldM}.
This means that an agent can only play with the dynamics of the environment to pretrain itself before being assigned with a specific task (i.e., having access to extrinsic rewards).
Therefore, we further investigate the first way to integrate mega-reward with extrinsic rewards (i.e., using mega-reward to pretrain the agent) and compare the pretrained agent with that in the state-of-the-art world model \cite{Ha2018WorldM},
as well as two state-of-the-art methods of unsupervised representation learning for RL: MOREL \cite{goel2018unsupervised} and OOMDP \cite{diuk2008object}.

\begin{figure}
	\centering
	\includegraphics[width=1.0\columnwidth]{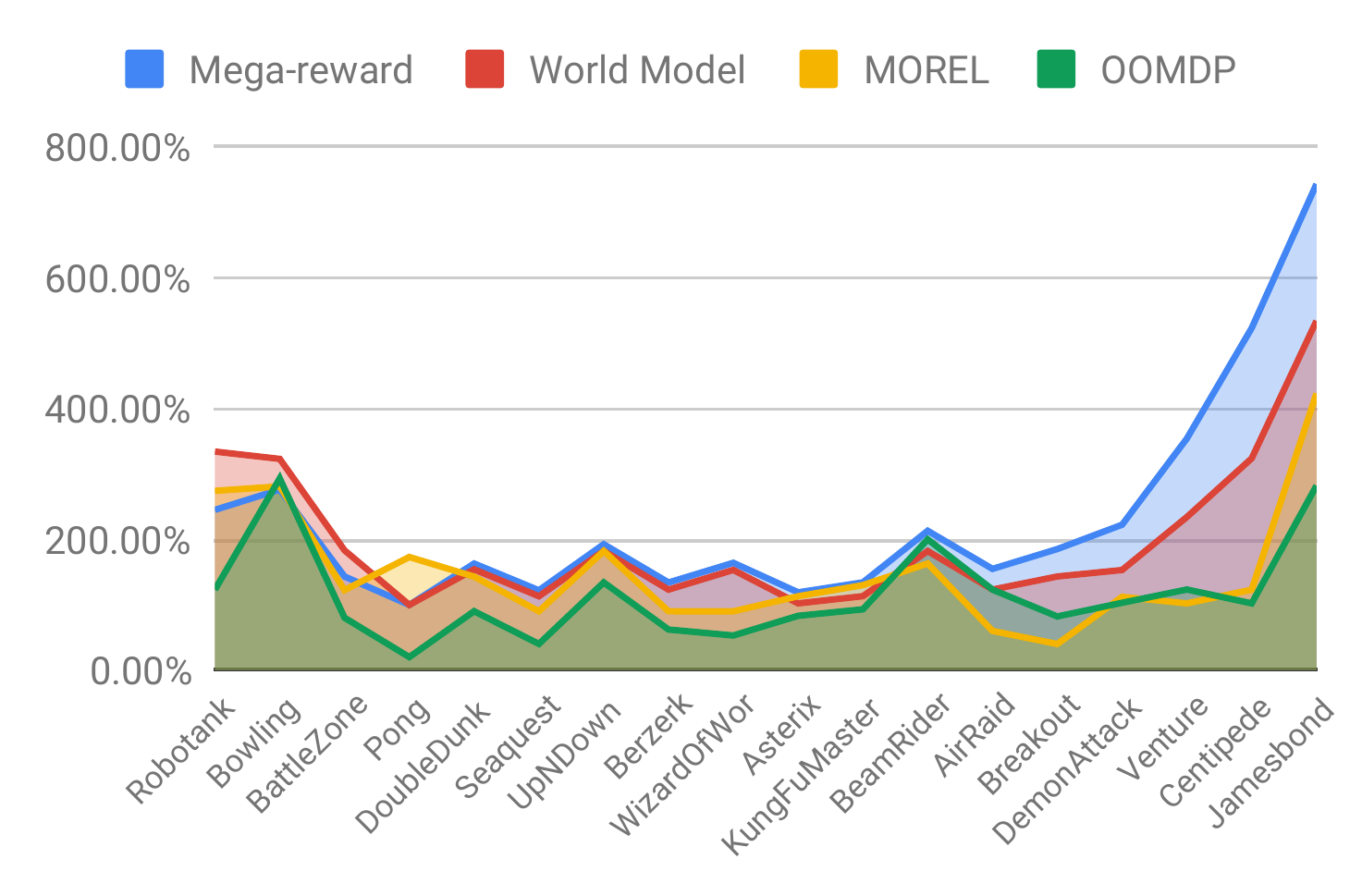}
	\caption{Relative improvements of the score when pretrained with mega-reward and world model (\textit{Delta} $=$ \textit{Mega-reward}-\textit{World Model}).}
	\label{pretrain-result}
\end{figure}

The evaluation is based on a relative improvement of the score, which is formally defined as
\begin{equation}
S_{\textrm{Improve}} = \frac{S_{\textrm{Pretrain}}-S_{\textrm{Random}}}{S_{\textrm{Scratch}}-S_{\textrm{Random}}} \times 100\%\,,
\end{equation}
where $S_{\textrm{Pretrain}}$ is the score after 20M steps with the first 10M steps pretrained without access to extrinsic rewards, and $S_{\textrm{Scratch}}$ is the score after 10M steps of training from scratch.
In 14, 15, and 17 out of 18 games (see Fig. \ref{pretrain-result}), pretraining using mega-reward achieves more relative improvements than pretraining using the world model, MOREL and OOMDP, respectively.
This shows that mega-reward is also very helpful for agents to achieve a superior performance when used in a domain with extrinsic rewards.

\subsection{Attention with Mega-Reward}
\label{attention-with-mega-reward}

Furthermore, ``{\em noisy TV}\/" is a long-standing open problem in novelty-driven approaches \cite{burda2018large,burda2018exploration}; it means that if there is a TV in the state that displays randomly generated noise at every step, the novelty-driven agent will find that watching at the noisy TV produces great interest.
A possible way to solve this problem is to have an attention mask to remove the state changes that are irrelevant to the agent, and we believe the accumulated latent control map $g^{h,w}_{t}$ can be used as such an attention mask.
Specifically, we estimate a running mean for each grid in $g^{h,w}_{t}$, which is then used to binarize $g^{h,w}_{t}$.
The binarized $g^{h,w}_{t}$ is used to mask the state used in the state-of-the-art novelty-driven work, \emph{RND}~\cite{burda2018exploration}, making RND generate novelty scores only related to the agent's control (both direct or latent).
There are two additional baselines, ADM \cite{choi2018contingency} and MOREL \cite{goel2018unsupervised} that also generate segmentation masks, which can be used to mask the state in RND.
Thus, we compare $g^{h,w}_{t}$ in our mega-reward with these baselines in terms of generating better masks to address the ``{\em noisy TV}\/" problem.

\begin{figure}
	\centering
	\includegraphics[width=1.0\columnwidth]{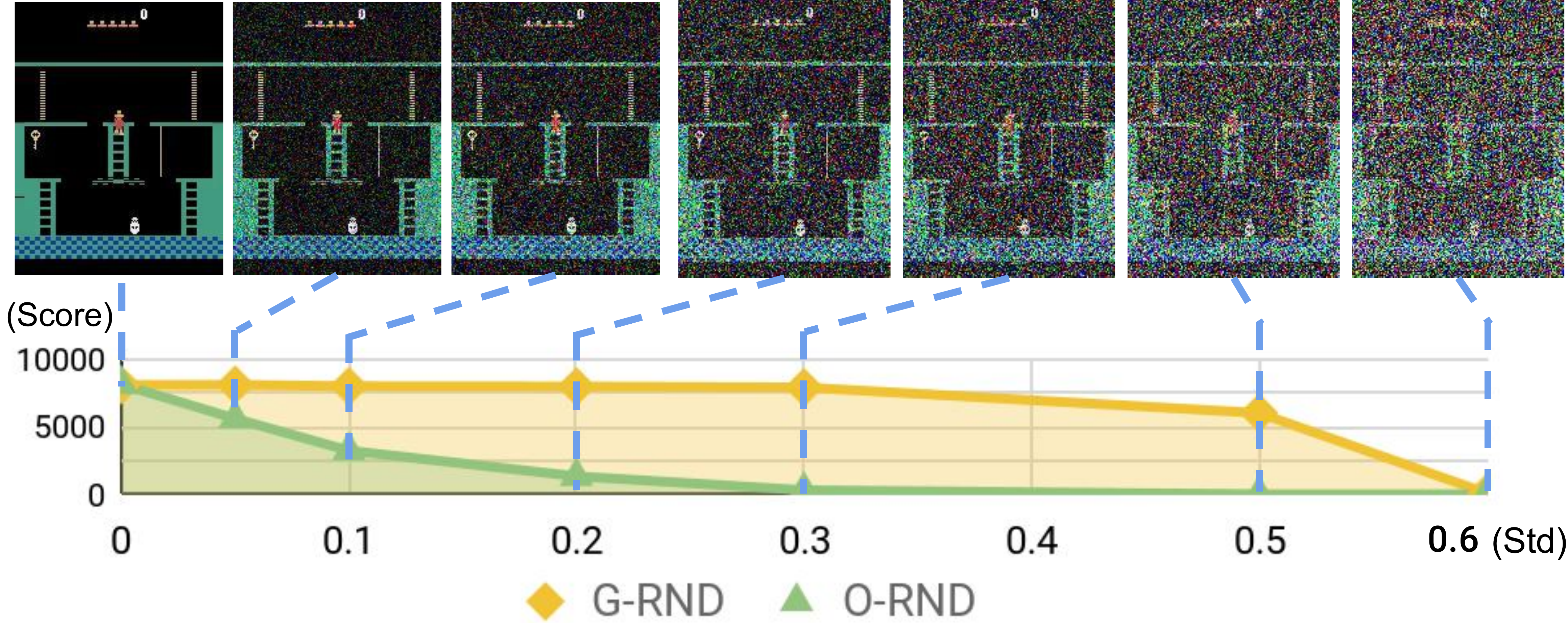}
	\caption{Comparing G-RND with O-RND over \textit{MontezumaRevenge} of different noises.}
	\label{attention-result}
\end{figure}

Experiments are first conducted on \textit{MontezumaRevenge}, following the same settings as in~\cite{burda2018exploration}.
Fig.~\ref{attention-result} shows the performance of the original RND (O-RND) and $g^{h,w}_{t}$-masked RND (G-RND) with different degrees of noise (measured by the STD of the normal noise).
The result shows that as the noise degree increases, the performance score of RND decreases catastrophically, while the performance drop of G-RND is marginal until the noise is so strong (STD $=$ 0.6) that it ruins the state representation.
Fig.~\ref{attention-result-all} shows the performance of 
O-RND and RND masked with different baselines (G-RND for $g^{h,w}_{t}$-masked RND, A-RND for ADM-masked RND, and M-RND for MOREL-masked RND) over 6 hard exploration Atari games with STD of $0.3$.
Results show that G-RND outperforms all other baselines, which means that $g^{h,w}_{t}$ generated in our mega-reward is the best mask to address the ``{\em noisy TV}\/" problem.
This further supports our conclusion that mega-reward can also achieve a superior performance when it is used together with extrinsic rewards.

\begin{table}
	\caption{Comparing G-RND with O-RND and other baselines over 6 hard exploration Atari games with STD of $0.3$.}
	\medskip
	\resizebox{\columnwidth}{!}{
		\begin{tabular}{lcccc}
			Game                  & \textrm{G-RND} & \textrm{A-RND} & \textrm{M-RND} & \textrm{O-RND} \\
			\midrule
			\textit{MontezumaRevenge}    & \textbf{7934}                        & 7138                       & 2385            & 313                       \\
      \textit{Gravitar}      &
      \textbf{3552}                       & 3485                       & 1634         & 1323                          \\
      \textit{Pitfall}      & -15.48                       & -14.43                       & \textbf{-13.64}         & -14.23                          \\
      \textit{PrivateEye}      & 7273                       & \textbf{7347}                       & 6128         & 2132                          \\
      \textit{Solaris}      & \textbf{3045}                       & 2857                       & 2253         & 2232                          \\
			\textit{Venture}      & 1700                       & \textbf{1701}                       & 1599         & 1572                          \\
		\end{tabular}
	}
	\label{attention-result-all}
\end{table}

\subsection{Failure Cases}
\label{failure-cases}

Some failure cases of mega-reward are also noticed. We find that mega-reward works well on most games with a meshing	 size of $4 \times 4$; however, some of the games with extremely small or big entities may fail with this size.
In addition, mega-reward also fails when the game terminates with a few seconds of flashing screen, because this will make the agent mistakenly believe that killing itself will flash the screen, which seems like having control on all entities for the agent.
Another failure case is that when the camera can be moved by the agent, such as in the game \textit{Pitfall} in Table~\ref{attention-result-all}.
The experiment's first step, i.e., modeling direct control $\alpha(s^{h,w}_{t},a_{t-1})$ via Eqs.~\eqref{adm-1} to 
\eqref{adm-4} fails, as all grids are under direct control when the agent moves its camera.
One possible solution for above failures is extracting the entities from the states using semantic segmentation~\cite{goel2018unsupervised}, then applying our method on the semantically segmented entities instead of each grid.

\section{Related Work}
We now discuss related works on intrinsic rewards. Further related work on contingency awareness, empowerment, variational intrinsic control, and relation-based networks is presented in the extended paper \cite{supplementary_material}.

Intrinsic rewards \cite{oudeyer2009intrinsic} are the rewards generated by the agent itself, in contrast to extrinsic rewards, which are provided by the environment.
Most  previous work on intrinsic rewards is based on the general idea of ``novelty-drivenness'', i.e., higher intrinsic rewards are given to states that occur relatively rarely in the history of an agent.
The general idea is also called ``surprise" or ``curiosity''.
Based on how to measure the novelty of a state, there are two classes of methods:  count-based methods \cite{bellemare2016unifying,martin2017count,ostrovski2017count,tang2017exploration} and prediction-error-based methods \cite{achiam2017surprise,pathak2017curiosity,burda2018large,burda2018exploration}.
Another popular idea to generate intrinsic rewards is ``difference-drivenness'', meaning that higher intrinsic rewards are given to the states that are different from the resulting states of other subpolicies \cite{florensa2017stochastic,SWLXX-AAAI-2019}.
To evaluate intrinsic rewards, intrinsically-motivated play has been adopted in several state-of-the-art works.
However, it may be an ill-defined problem, i.e., if we flip the extrinsic rewards, the agent only trained by the intrinsic rewards is likely to perform worse than a random agent in terms of the flipped extrinsic rewards.
Discarding the possible bug in defining the problem, intrinsically-motivated play indeed helps in many scenarios, such as pretraining, improving exploration, as well as understanding human intelligence.

\section{Summary}

In this work, we proposed a novel and powerful intrinsic reward, called mega-reward, to maximize the control over given entities in a given environment.
To our knowledge, mega-reward is the first approach that achieves the same level of performance as professional human players in intrinsically-motivated play. To formalize mega-reward, we proposed a relational transition model to bridge the gap between direct and latent control. Extensive experimental studies are conducted to show the superior performance of mega-reward in both intrinsically-motivated play and real-world scenarios with extrinsic rewards.

\subsubsection{Acknowledgments.}
This work was supported by the China Scholarship Council under the State Scholarship Fund, by the Graduate Travel and Special Project Grants from the Somerville College of the University of Oxford, by the Alan Turing Institute under the UK EPSRC grant EP/N510129/1, by the AXA Reseach Fund, by the National Natural Science Foundation of China under the grants 61906063, 61876013, and 61922009, by the Natural Science Foundation of Tianjin under the grant 19JCQNJC00400, by the ``100 Talents Plan’’ of Hebei Province, and by the Yuanguang Scholar Fund of Hebei University of Technology.

\clearpage
\onecolumn

\section{Related Work: Contingency Awareness, Empowerment, Variational Intrinsic Control and Relation-based Networks}

The concept of \textit{contingency awareness} originally comes from psychology  \cite{watson1966development,baeyens1990contingency}, where infants are proved to be aware that the entities in the state are potentially related to their actions.
The idea was first introduced into AI  by \cite{bellemare2012investigating}.
More recently, the discovery of \textit{grid cells} \cite{moser2015place}, a neuroscience finding that supports the psychology concept of contingency awareness, triggered the interest of applying grid cells in AI agents \cite{banino2018vector,whittington2018generalisation}.
Another popular idea developed from contingency awareness is the one of \textit{inverse models}, which are used to learn representations that contain the necessary information about action-related changes in states \cite{pathak2017curiosity}, or generate attention masks about which part of the states is action-related \cite{choi2018contingency}.
Other ideas following contingency awareness include controllable feature learning \cite{forestier2017intrinsically,laversanne2018curiosity}, tool use discovery \cite{forestier2016modular} and sensing guidance \cite{klyubin2005all,klyubin2005empowerment,klyubin2008keep}.
However, we formalize contingency awareness into a powerful intrinsic reward (mega-reward) for human-level intrinsically-motivated play.
Besides,  existing works are only capable of figuring out what is under the agent's direct control, while we build the awareness of latent control and show that the awareness of latent control is the key to achieving a powerful intrinsic reward.

The idea of ``having more control over the environment'' is also mentioned in \textit{empowerment}  \cite{klyubin2005all,klyubin2008keep}, which,
however, is commonly based on mutual information between the actions and the entire state \cite{mohamed2015variational,montufar2016information}, the latter of which evolves into stochasticity-drivenness \cite{florensa2017stochastic}.
While our megalomania-drivenness is based on identifying how actions are latently related to each entity in the state, which evolves from contingency awareness \cite{watson1966development}.
Thus, ``megalomania-drivenness''  is different from ``empowerment''.

Another body of related work that is Variational Intrinsic Control \cite{gregor2016variational}, and its followup works, e.g., DIAYN \cite{eysenbach2018diversity}, DISCERN \cite{warde2018unsupervised}, VISR \cite{hansen2019fast}.
They extend empowerment from actions to a fixed-length action sequence, i.e., an option.
However, mega-reward does not need the prior of this fixed length, i.e., the control effect of action can delay into the future for unknown steps, which is a more practical setting.

A part of RTM, $\Phi$, is similar to \textit{relation-based networks} \cite{battaglia2016interaction,santoro2017simple,watters2017visual,wu2017adversarial,van2018relational,battaglia2018relational}, which have recently been applied to predict temporal transitions \cite{watters2017visual} and to learn representations in RL \cite{zambaldi2018deep}.
However, relation-based networks do not explicitly model mega-reward's $\alpha(s^{h,w}_{t}, s^{h',w'}_{t-1})$, while RTM model it with $\Gamma$.
Thus, RTM are defined and trained in a different way.

\section{Ablation Study of Grid Size}

\begin{figure}
	\centering
	\centerline{\includegraphics[width=0.7\textwidth]{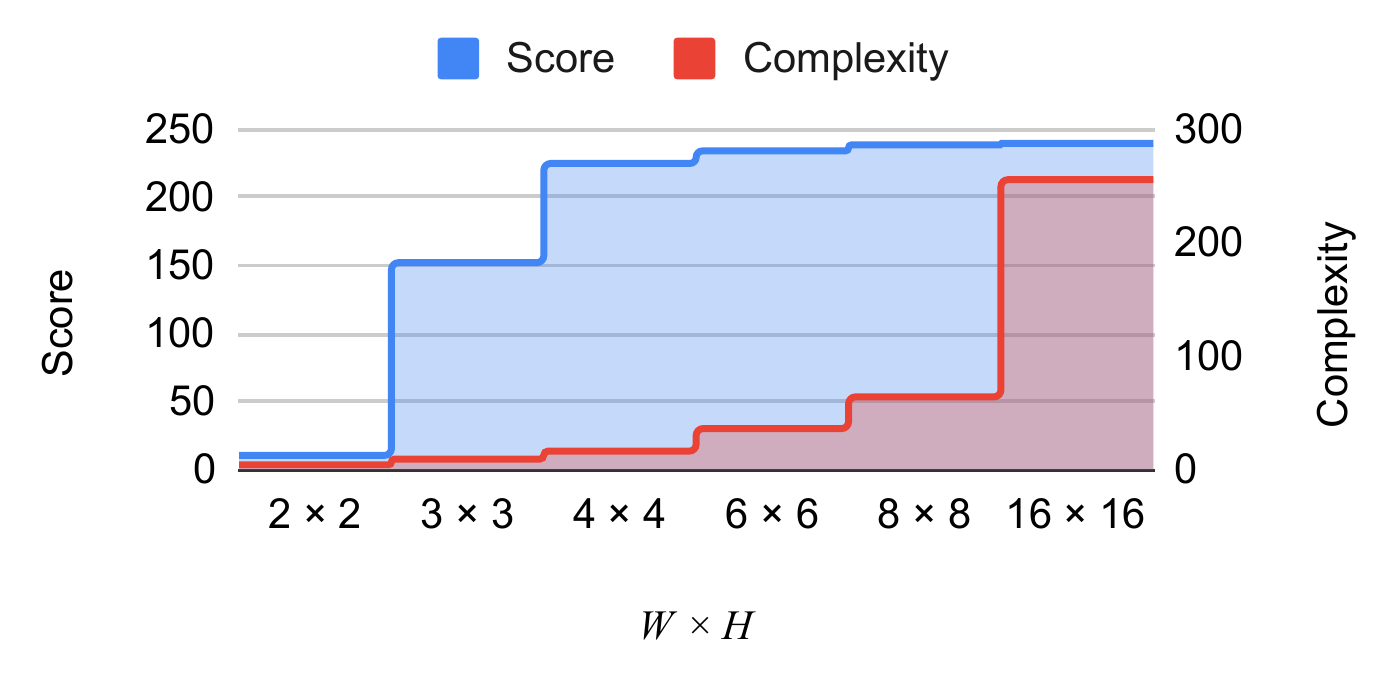}}
	\caption{Ablation Study of $H \times W$.}
	\label{ablation-hw}
\end{figure}

\textcolor{black}{
$H \times W=4 \times 4$ is a trade-off between efficiency and accuracy.
We conducted an ablation study on value settings of $H \times W$ on the game \textit{Breakout} in Fig. \ref{ablation-hw}, showing that $4 \times 4$ is sufficient for Atari to achieve a reasonable accuracy, while having the best efficiency (low in complexity).
}

\newpage
\section{Neural Network Details}

The details of the network architectures  to model $\Phi$ and  $\Gamma$ are shown in Tables \ref{network_phi} and \ref{network_gamma}, respectively.
A fully connected layer is denoted  FC, and a flatten layer is denoted  Flatten.
We use leaky rectified linear units (denoted LeakyRelu) \cite{maas2013rectifier} with
leaky rate $0.01$ as the nonlinearity applied to all the hidden layers in our network.
Batch normalization \cite{ioffe2015batch} (denoted BatchNorm) is applied after hidden convolutional layers (denoted  Conv).
To model $\Phi$ and $\Gamma$, the integration of the three inputs is accomplished by approximated multiplicative interaction \cite{oh2015action} (the \emph{dot-multiplication}  in Tables \ref{network_phi} and  \ref{network_gamma}), so that any predictions made by $\Phi$ or  $\Gamma$ are conditioned on the three inputs together.
Deconvolutional layers (denoted  DeConv) \cite{zeiler2011adaptive} in $\Phi$ are applied for predicting relational transitions.

\medskip 
\begin{table}
	\centering
	\resizebox{1.0\columnwidth}{!}{
	\begin{tabular}{c|c|c}
		\midrule
		\textbf{Input 1}: $\mathbf{S}^{h',w'}_{t-1}$  &  \textbf{Input 2}: $a_{t-1}$, as one-hot vector &  \textbf{Input 3}: $(h-h',w-w')$, as one-hot vector\\
        \midrule
		\textbf{Conv}: kernel size $5\times5$, number of features 16, stride 2 & & \\
		\textbf{BatchNorm} & & \\
        \textbf{LeakyRelu} & & \\
        \textbf{Conv}: kernel size $4\times4$, number of features 32, stride 1 & & \\
		\textbf{BatchNorm} & \textbf{FC}: number of features 1024 & \textbf{FC}: number of features 1024 \\
        \textbf{LeakyRelu} & & \\
        \textbf{Flatten}: $32\times6\times6$ is flattened to $1152$ & & \\
        \textbf{FC}: number of features 1024 & &  \\
        \textbf{BatchNorm} & &  \\
        \textbf{LeakyRelu} & &  \\
		\midrule
		\multicolumn{3}{c}{\textbf{Dot-multiply}}\\
        \midrule
        \multicolumn{3}{c}{\textbf{FC}: number of features 1152}\\
        \multicolumn{3}{c}{\textbf{BatchNorm}}\\
        \multicolumn{3}{c}{\textbf{LeakyRelu}}\\
        \multicolumn{3}{c}{\textbf{Reshape}: $1152$ is reshaped to $32\times6\times6$}\\
        \multicolumn{3}{c}{\textbf{DeConv}: kernel size $4\times4$, number of features 16, stride 1}\\
        \multicolumn{3}{c}{\textbf{BatchNorm}}\\
        \multicolumn{3}{c}{\textbf{LeakyRelu}}\\
        \multicolumn{3}{c}{\textbf{DeConv}: kernel size $5\times5$, number of features 1, stride 2}\\
        \multicolumn{3}{c}{\textbf{Tanh}}\\
		\midrule
        \multicolumn{3}{c}{\textbf{Output}: $\Phi \left(\left[\mathbf{S}^{h',w'}_{t-1},a_{t-1},h-h',w-w'\right]\right)$ } \\
		\midrule
	\end{tabular}
	}
    \caption{Network architecture of $\Phi$.}
	\label{network_phi}
\end{table}

\begin{table}
	\centering
	\resizebox{1.0\columnwidth}{!}{
	\begin{tabular}{c|c|c}
		\midrule
		\textbf{Input 1}: $\mathbf{S}^{h',w'}_{t-1}$  &  \textbf{Input 2}: $a_{t-1}$, as one-hot vector &  \textbf{Input 3}:  $(h-h',w-w')$, as one-hot vector\\
        \midrule
		\textbf{Conv}: kernel size $5\times5$, number of features 16, stride 2 & & \\
		\textbf{BatchNorm} & & \\
        \textbf{LeakyRelu} & & \\
        \textbf{Conv}: kernel size $4\times4$, number of features 32, stride 1 & & \\
		\textbf{BatchNorm} & \textbf{FC}: number of features 1024 & \textbf{FC}: number of features 1024 \\
        \textbf{LeakyRelu} & & \\
        \textbf{Flatten}: $32\times6\times6$ is flattened to $1152$ & & \\
        \textbf{FC}: number of features 1024 & &  \\
        \textbf{BatchNorm} & &  \\
        \textbf{LeakyRelu} & &  \\
		\midrule
		\multicolumn{3}{c}{\textbf{Dot-multiply}}\\
        \midrule
        \multicolumn{3}{c}{\textbf{FC}: number of features 512}\\
        \multicolumn{3}{c}{\textbf{BatchNorm}}\\
        \multicolumn{3}{c}{\textbf{LeakyRelu}}\\
        \multicolumn{3}{c}{\textbf{Tanh}}\\
        \multicolumn{3}{c}{\textbf{FC}: number of features 1}\\
		\midrule
        \multicolumn{3}{c}{\textbf{Output}: $\bar{\gamma}^{h,w \leftarrow h',w'}$ } \\
		\midrule
	\end{tabular}
	}
	\caption{Network architecture of $\Gamma$.}
	\label{network_gamma}
\end{table}

\newpage 
\section{Results for More Baselines}

\textcolor{black}{
Results for more baselines of intrinsically-motivated play can be found in Table \ref{raw-score-additional}.
}

\begin{table}
	\centering
	\begin{tabular}{lc}
		Game                  & \textit{Count-based} \cite{martin2017count} \\
		\midrule
		\textit{Seaquest}     & 234.2      \\
		\textit{Bowling}      & 100.2    \\
		\textit{Venture}      & 0            \\
		\textit{WizardOfWor}  & 465.2             \\
		\textit{Asterix}      & 432.1               \\
		\textit{Robotank}     & 1.953                \\
		\textit{BeamRider}    & 582.1           \\
		\textit{BattleZone}   & 2459               \\
		\textit{KungFuMaster} & 235                   \\
		\textit{Centipede}    & 1263           \\
		\textit{Pong}         & -15.36              \\
		\textit{AirRaid}      & 868.3             \\
		\textit{DoubleDunk}   & -20.51          \\
		\textit{DemonAttack}  & 36.42        \\
		\textit{Berzerk}      & 356.9          \\
		\textit{Breakout}     & 126.8         \\
		\textit{Jamesbond}    & 1893     \\
		\textit{UpNDown}      & 9452
	\end{tabular}
	\vspace{0.5em}
	\caption{Results for more baselines of intrinsically-motivated play.}
	\label{raw-score-additional}
\end{table}
\section{Running Time}

Table \ref{learning_speed} shows the learning speed on the game Seaquest of our approach against baselines and benchmarks on running time (hours).

\begin{table}
	\centering
	\begin{tabular}{lcccccccc}
		Game                  & \textit{Emp} & \textit{Cur} & \textit{RND} & \textit{Sto} & \textit{Div} &  \textit{Dir} & \textit{Meg} & \textit{Ex-PPO} \\
		\midrule
		\textit{Seaquest}    & 14.24                       & 15.87                       & 17.62            & 12.96                             &  19.66                         & 21.32                  & 34.22                 & \textbf{5.126}                 \\
	\end{tabular}
\smallskip
	\caption{Comparison of mega-reward against baselines and benchmarks on running time (hours);  conducted on a server with i7 CPU (16 cores), and one Nvidia GTX 1080Ti GPU. Each method ran for 80M frames.}
	\label{learning_speed}
\end{table}
\section{Ablation Study of Components in Mega-Reward}

\begin{figure}
	\centering
	\centerline{\includegraphics[width=0.7\textwidth]{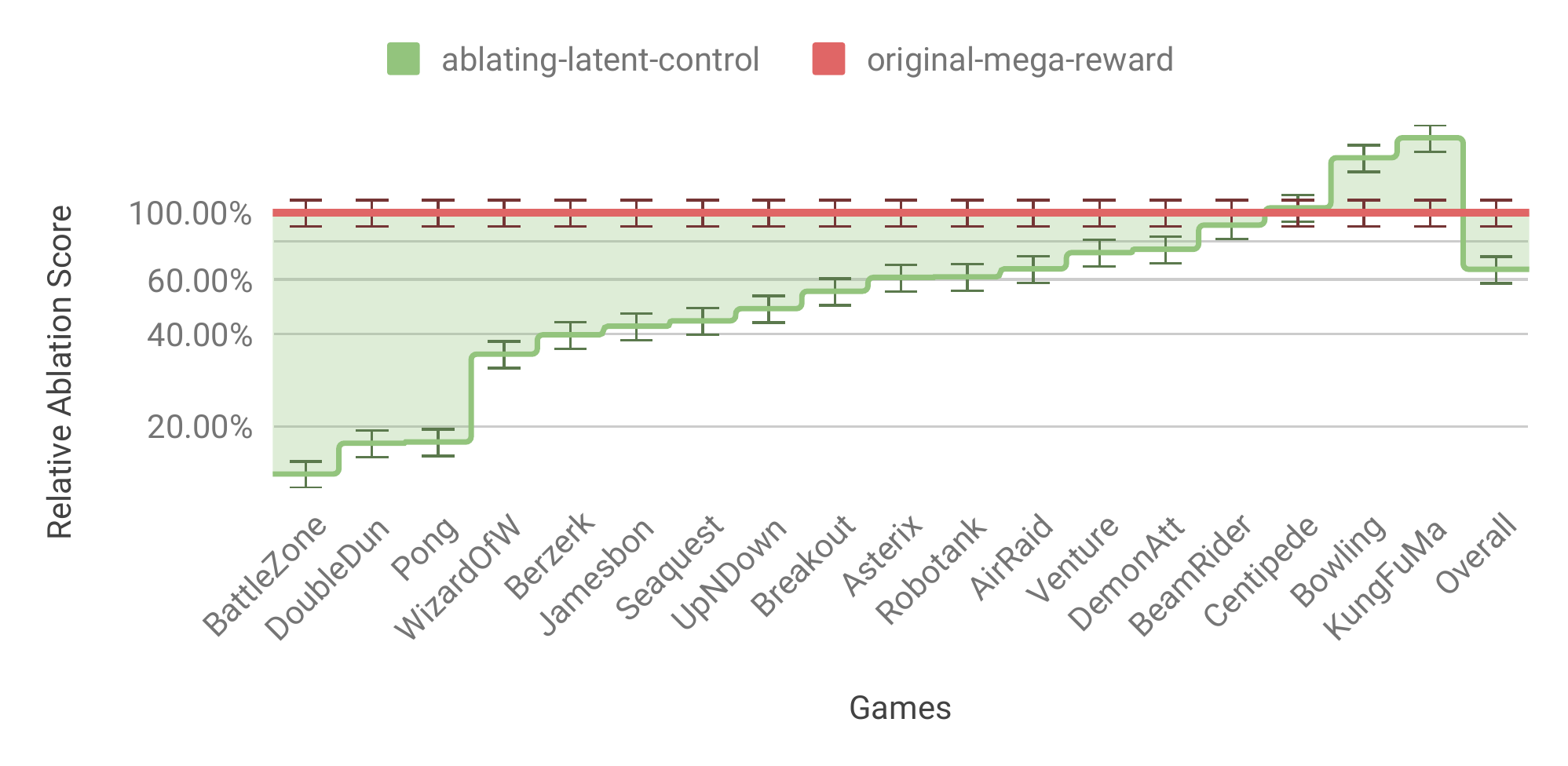}}
	\caption{Ablation study of components in mega-reward.}
	\label{ablation-components-1}
\end{figure}

\begin{figure}
	\centering
	\centerline{\includegraphics[width=0.7\textwidth]{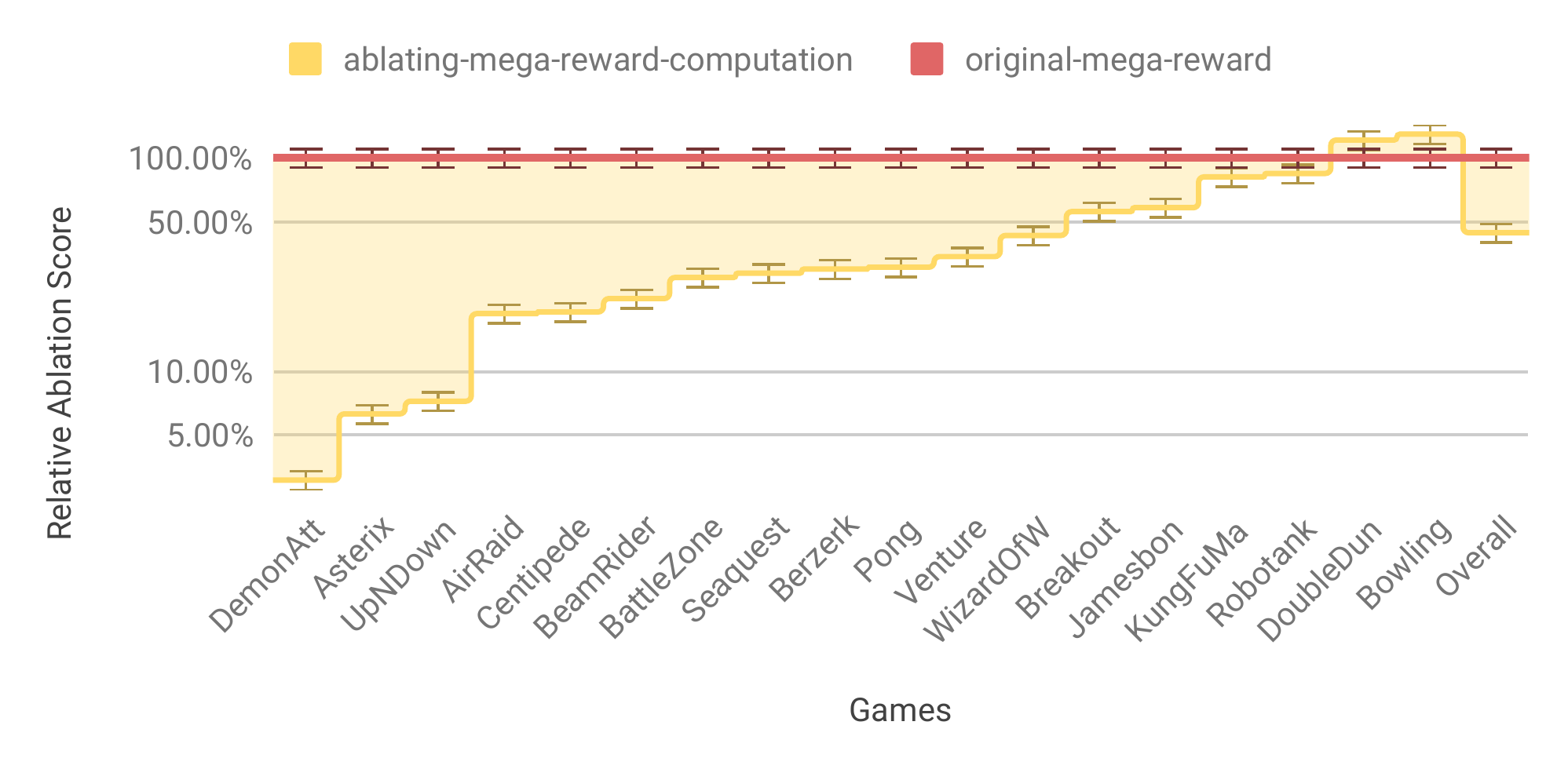}}
	\caption{Ablation study of components in mega-reward.}
	\label{ablation-components-2}
\end{figure}

\textcolor{black}{
The proposed mega-reward has three components:
\begin{itemize}
	\item[(1)] ADM \cite{choi2018contingency}, which produces quantification of direct control $\alpha(s^{h,w}_{t},a_{t-1})$;
	\item[(2)] RTM, which produces quantification of latent control $g^{h,w}_{t}$;
	\item[(3)] Eqs.\ (10) to (11) in the main paper, which compute mega-reward.
\end{itemize}
Figs.~\ref{ablation-components-1} and \ref{ablation-components-2} show the results of ablating different components.
Specifically, in Figs.~\ref{ablation-components-1}, we observe that when applying (3) over (1), which means apply Eqs. (10) to (11) to $\alpha(s^{h,w}_{t},a_{t-1})$ instead of $g^{h,w}_{t}$ to compute the intrinsic reward, the score drops by 34.82\% overall comparing to the \textit{original-mega-reward}.
This ablation is called \textit{ablating-latent-control} and the result indicates the necessity of modeling latent control; when replacing (3) with count-based novelty (this means applying a count-based novelty estimation on $g^{h,w}_{t}$, so that to produce a novelty quantification of $g^{h,w}_{t}$, which can be used as an intrinsic reward), i.e., \textit{ablating-mega-reward-computation}, the score drops by 55.48\% overall, compared to the \textit{original-mega-reward}, indicating the effectiveness of Eqs. (10) to (11) in computing a good intrinsic reward.
Note that the relative ablation score in Figs. \ref{ablation-components-1} and \ref{ablation-components-2} is given by:
\begin{equation}
\label{relative-ablation}
\textrm{Relative Ablation Score} = \frac{\textrm{Ablation Score}-\textrm{Random Score}}{\textrm{Original Mega-reward Score}-\textrm{Random Score}} \times 100\%.
\end{equation}
}

\newpage 
\section{Proof of Lemmas}

\newtheorem{lemma}{Lemma}
\begin{lemma}
	\label{lemma-1}
	\begin{equation}
    g^{h,w}_{t} = \rho \sum_{\substack{h' \in H, w' \in W}} \alpha(s^{h,w}_{t}, s^{h',w'}_{t-1}) g^{h',w'}_{t-1} + \alpha(s^{h,w}_{t},a_{t-1})
    \end{equation}
\end{lemma}

\noindent \textbf{Proof of Lemma~\ref{lemma-1}:}

{\begin{align}
&g^{h,w}_{t} =\!\!\! \sum_{n\in\left[ 1,t \right]} \rho^{n-1} \alpha(s^{h,w}_{t},a_{t-n}) =\!\!\! \sum_{n\in\left[ 2,t \right]} \rho^{n-1} \alpha(s^{h,w}_{t},a_{t-n}) + \alpha(s^{h,w}_{t},a_{t-1}) \nonumber \\
&= \sum_{n \in [2,t]} \rho^{n-1} \overbrace{\sum_{\substack{h' \in H, w' \in W}} \alpha(s^{h,w}_{t}, s^{h',w'}_{t-1}) \alpha(s^{h',w'}_{t-1},a_{t-n}) }^{=\ \alpha(s^{h,w}_{t},a_{t-n}) \textrm{ according to Eq. (5) in paper}} + \alpha(s^{h,w}_{t},a_{t-1}) \nonumber \\
&= \rho \sum_{n \in [2,t]} \rho^{n-2} \sum_{\substack{h' \in H, w' \in W}} \alpha(s^{h,w}_{t}, s^{h',w'}_{t-1}) \alpha(s^{h',w'}_{t-1},a_{t-n}) + \alpha(s^{h,w}_{t},a_{t-1}) \nonumber \\
&= \rho \sum_{\substack{h' \in H, w' \in W}} \alpha(s^{h,w}_{t}, s^{h',w'}_{t-1}) \underbrace {\sum_{n \in [2,t]} \rho^{n-2}  \alpha(s^{h',w'}_{t-1},a_{t-n})}_{=\ g^{h' , w'}_{t-1} \textrm{ according to Eq. (9) in paper}} + \alpha(s^{h,w}_{t},a_{t-1}) \nonumber \\
&= \rho \sum_{\substack{h' \in H, w' \in W}} \alpha(s^{h,w}_{t}, s^{h',w'}_{t-1}) g^{h',w'}_{t-1} + \alpha(s^{h,w}_{t},a_{t-1}) \,.
\end{align}}%

\section{Benefits of mega-reward over other intrinsically motivated approaches}

The baselines measuring novelty ($Cur$, $RND$) failed when attracted by ``interesting'' things not agent-controllable. The ones of HRL ($Sto$, $Div$) failed, as they only learn useful subpolicies; the master policy still acts randomly. Others ($Emp$, $Dir$) failed, as they cannot efficiently model latent control. But mega-reward can overcome all these failures by efficiently modeling latent control.

\clearpage
\twocolumn

\bibliographystyle{aaai}
\bibliography{1506-yuhang}

\end{document}